\pdfoutput=1

\documentclass[11pt,table]{article}
\usepackage[final]{acl}

\usepackage{times}
\usepackage{latexsym}
\usepackage[T1]{fontenc}
\usepackage[utf8]{inputenc}

\usepackage{amsfonts}
\usepackage{amsmath}
\usepackage{amssymb}        
\usepackage{bold-extra}     
\usepackage{bm}             

\usepackage{microtype}
\usepackage{graphicx}       
\usepackage{multirow}       
\usepackage{booktabs}       
\usepackage{enumitem}       
\usepackage{subcaption}     

\usepackage[ruled,linesnumbered,noend]{algorithm2e}
\usepackage[hang,flushmargin]{footmisc}  
\usepackage{xcolor}

\usepackage{pgf-pie}
\usepackage{algpseudocode} 
\usepackage{svg}
\usepackage{cuted}
\usepackage{url}

\newcommand{\LN}{\linebreak\noindent}    

\usepackage{enumitem}
\usepackage{cuted}
\setenumerate[1]{leftmargin=*}    
\setitemize[1]{leftmargin=*}      

\title{Aligning Speakers: Evaluating and Visualizing Text-based Diarization\\ Using Efficient Multiple Sequence Alignment (Extended Version)}

\author{Chen Gong \\
  Computer Science \\
  Emory University \\
  Atlanta, GA 30322 USA \\
  \texttt{\small chen.gong@emory.edu} \\\And
  Peilin Wu \\
  Computer Science \\
  Emory University \\
  Atlanta, GA 30322 USA \\
  \texttt{\small peilin.wu@emory.edu} \\\And
  Jinho D. Choi \\
  Computer Science \\
  Emory University \\
  Atlanta, GA 30322 USA \\
  \texttt{\small jinho.choi@emory.edu} \\}

\begin{document}
\maketitle

\begin{abstract}

This paper presents a novel evaluation approach to text-based speaker diarization (SD), tackling the limitations of traditional metrics that do not account for any contextual information in text.
Two new metrics are proposed, Text-based Diarization Error Rate and Diarization F1, which perform utterance- and word-level evaluations by aligning tokens in reference and hypothesis transcripts.
Our metrics encompass more types of errors compared to existing ones, allowing us to make a more comprehensive analysis in SD.
To align tokens, a multiple sequence alignment algorithm is introduced that supports multiple sequences in the reference while handling high-dimensional alignment to the hypothesis using dynamic programming.
Our work is packaged into two tools, align4d providing an API for our alignment algorithm and TranscribeView for visualizing and evaluating SD errors, which can greatly aid in the creation of high-quality data, fostering the advancement of dialogue systems.

\end{abstract}

\section{Introduction}
\label{sec:introduction}

The rise of data-driven dialogue systems, such as BlenderBot \cite{blendorbot-3} and ChatGPT\footnote{\url{https://chat.openai.com}}, powered by large language models \cite{NEURIPS2020_1457c0d6,lewis-etal-2020-bart,t5}, has generated significant interest across various groups.
Conversational AI has emerged as a central focus for numerous organizations, presenting a wealth of potential applications.
Many institutes have started utilizing recordings of human-to-human dialogues collected over the years to develop dialogue models.
However, the majority of these recordings were not intended for data-driven model development originally, resulting in low-quality audio with prominent background noise. 
This poses inevitable challenges for automatic speech recognition (ASR) systems, while the lack of dedicated channels for individual\LN speakers necessitates the use of robust speaker diarization (SD) techniques.

\noindent SD is a speech processing task to identify speakers of audio segments extracted from a conversation involving two or more speakers \cite{Park-etal-2021-speaker-diarization-review}.
Despite the excellent performance of ASR models for translating audio into text without recognizing speakers \cite{NEURIPS2020_92d1e1eb,conformer,whisper}, unstable SD has a detrimental effect on developing an accurate dialogue system as models trained on such data would fail to learn the distinct languages and characteristics of individual speakers.
Thus, it is crucial to thoroughly assess the ASR and SD performance to generate high-quality transcripts.
Nonetheless, there is no comprehensive platform available that allows for the simultaneous evaluation of both ASR and SD errors.

\vspace{-1em}
\begin{figure}[htbp!]
\centering
\includegraphics[width=\columnwidth]{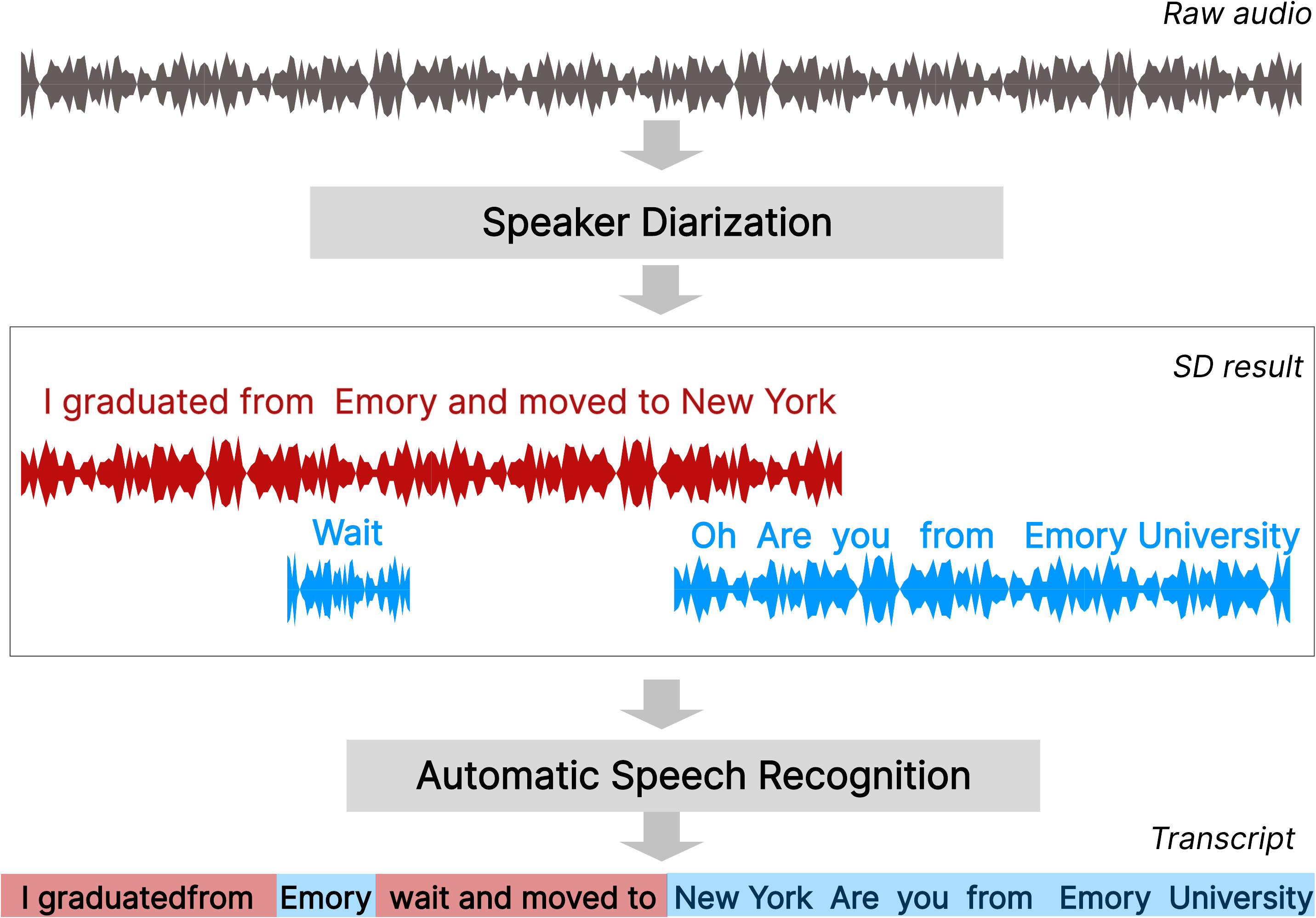}
\caption{An example illustrating speaker diarization errors introduced during automatic speech recognition.}
\label{fig:SD-ASR}
\vspace{-1em}
\end{figure}

\noindent Traditionally, SD performance has been evaluated on audio segments by testing the system's ability to recognize the number of speakers in each segment.
However, these segments are often uniformly split from an audio stream, disregarding speaker context.
A more insightful analysis can be made by correctly aligning tokens between reference and hypothesis transcripts and directly evaluating SD performance on the transcripts, where utterances are segmented based on speaker turns.
This work carefully revisits traditional ASR/SD evaluation metrics (Section~\ref{sec:related-work}) and compares them with our new approach to verify its effectiveness (Section~\ref{sec:experiments}).
Our contributions are:

\begin{enumerate}
\setlength\itemsep{0em}
\item An efficient multiple sequence alignment algorithm that maps tokens between reference and hypothesis transcripts (Sections~\ref{sec:align4d}).

\item Two metrics for evaluating the task of text-based speaker diarization (Section~\ref{sec:text-based-sd-evaluation}).

\item A CPython API for our alignment algorithm and a web-based visualization interface for analysis of ASR and SD errors (Section~\ref{sec:applications}).

\end{enumerate}

\section{Background}
\label{sec:related-work}

This section provides a brief overview of the most commonly used evaluation metrics for audio-based SD (Section~\ref{ssec:der}), text-based SD (Section~\ref{ssec:wder}), and ASR (Section~\ref{ssec:wer}), as well as their limitations.

\subsection{Diarization Error Rate}
\label{ssec:der}

For a set of audio segments $\mathcal{S}$, the SD performance is often tested using Diarization Error Rate (\texttt{DER}), which measures the proportion of time in an audio segment incorrectly attributed to a speaker or left unassigned \cite{DER}:
\begin{equation}
\label{eq:DER-full}
\small
\texttt{DER} = \frac{\sum_{\forall s \in \mathcal{S}}(dur(s) \cdot (\max(N_{r}(s), N_{h}(s))-N_{c}(s)))}{\sum_{\forall s \in \mathcal{S}} dur(s) \cdot N_{r}(s)}
\end{equation}

\noindent $dur(s)$ is the time duration of an audio segment $s$.
$N_{r}(s)$ and $N_{h}(s)$ are the numbers of speakers in $s$ given the reference (ground-truth) and hypothesis (system-generated) transcripts, respectively.
$N_{c}(s)$ is the number of correctly identified speakers in $s$.
For a more detailed analysis, \texttt{DER} can be decomposed into four types of diarization errors:

\paragraph{Speaker Error}\hspace{-1em} occurs when a segment is attributed to a wrong speaker:
\begin{equation}
\label{eq:DER-SE}
\small
\begin{split}
\mathcal{T} &= \{s : \forall s \in \mathcal{S}.\: N_{h}(s) = N_{r}(s)\} \\[0.3em]
E_{se} &= \frac{\sum_{\forall t \in \mathcal{T}} dur(t) \cdot (N_{*}(t) - N_{c}(t))}{\sum_{\forall t \in \mathcal{T}} dur(t) \cdot N_{r}(t)}
\end{split}
\end{equation}

\paragraph{False Alarm}\hspace{-1em} occurs when a non-speech segment (e.g., pause) is assigned to a speaker, or more speakers than actual ones are identified for a segment:
\begin{equation}
\label{eq:DER-FA}
\small
    \begin{split}
    \mathcal{T} &= \{s : \forall s \in \mathcal{S}.\: N_{h}(s) > N_{r}(s)\} \\[0.3em]
    E_{fa} & = \frac{\sum_{\forall t \in \mathcal{T}} dur(t)\cdot(N_{h}(t) - N_{r}(t))}{\sum_{\forall t \in \mathcal{T}} dur(t)\cdot N_{r}(t)}
    \end{split}
\end{equation}

\paragraph{Missed Speech}\hspace{-1em} occurs when the system misses to recognize a segment from a speaker, resulting in a gap in the speaker's transcript:
\begin{equation}
\label{eq:DER-MS}
\small
    \begin{split}
    \mathcal{R} &= \{s : \forall s \in \mathcal{S}.\: N_{h}(s) < N_{r}(s)\} \\[0.3em]
    E_{ms} & = \frac{\sum_{\forall r \in \mathcal{R}} dur(r)\cdot(N_{r}(r) - N_{h}(r))}{\sum_{\forall r \in \mathcal{R}} dur(r) \cdot N_{r}(r)}
    \end{split}
\end{equation}

\paragraph{Overlapping Speech}\hspace{-1em} occurs when multiple speakers speak at the same time and the system fails to recognize all speakers in a segment. In this case, $N_{h}(s) < N_{r}(s)$, and thus, it is included in $E_{ms}$.

Given this decomposition, \texttt{DER} can be reformulated as follows:
\begin{equation}
\label{eq:DER-short}
\small
    \texttt{DER} = E_{se} + E_{fa} + E_{ms}
\end{equation}

\subsection{Word-level Diarization Error Rate}
\label{ssec:wder}

Current state-of-the-art results in SD are achieved by jointly training ASR \& SD \cite{joint-asr-sds}, leading to the need for new evaluation metrics beyond traditional audio-based metrics such as \texttt{DER}. 
Thus, Word-level Diarization Error Rate (\texttt{WDER}) is proposed to evaluate the SD performance of such joint systems \cite{Park-etal-2021-speaker-diarization-review}.
Unlike \texttt{DER} that focuses only on time-based errors, \texttt{WDER} provides a more detailed evaluation of SD performance by considering the alignment of words and speakers in the transcriptions as follows: 
\begin{equation}
\label{eq:WDER}
\small
    \texttt{WDER} = \frac{U_{s} + O_{s}}{U + O}
\end{equation}
$U$ is the set of substitutions, where each substitution replaces the actual word with an incorrect one, and $O$ is the set of correctly recognized words.
$U_s$ and $O_s$ are the subsets of words in $U$ and $O$ respectively, whose speaker IDs are incorrectly identified.

It is important to note that \texttt{WDER} only takes into account the words aligned between the reference and hypothesis transcripts, $U$ and $O$ in Equation~\ref{eq:WDER}, such that it does not consider inserted and deleted words.
As a result, among the four types of errors in Section~\ref{ssec:der}, \texttt{WDER} only captures speaker errors; the other 3 types of errors, reflected in the deleted and inserted words, are not assessed by \texttt{WDER}.

\subsection{Word Error Rate}
\label{ssec:wer}

Word Error Rate (\texttt{WER}) is a commonly used metric for evaluating ASR performance \cite{WER}. It quantifies the similarity between the reference and hypothesis transcripts by counting the min-number of edit operations (insertions, deletions, and substitutions) required to transform the hypothesis into the reference and dividing it by the total number of words in the reference as follows:
\begin{equation}
\small
\texttt{WER} = \frac{\text{\#(insertions)} + \text{\#(deletions)} + \text{\#(substitutions)}}{\text{Total \# of Words in Reference}}
\end{equation}
While \texttt{WER} is widely adapted, it focuses solely on word-level errors and does not consider speaker information so that it cannot capture errors related to speaker identification or segmentation.
Therefore,\LN \texttt{WER} is inadequate for evaluating SD.

\section{Multiple Sequence Alignment}
\label{sec:align4d}

\begin{figure}[htbp!]
    \centering
    \includegraphics[width=0.9\columnwidth]{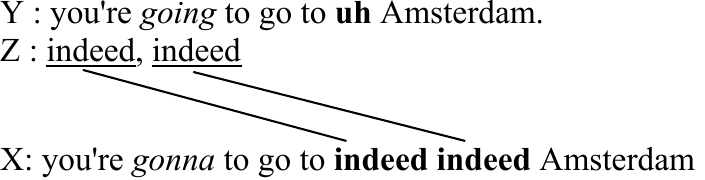}
    \caption{Examples of transcript errors, where the reference consists of multiple sequences.}
    \label{fig:transcript_errors}
\vspace{-0.5em}
\end{figure}

To evaluate text-based SD (Section~\ref{sec:text-based-sd-evaluation}), tokens in the\LN hypothesis transcript must be aligned with the most similar tokens in the reference transcript.
In Fig.~\ref{fig:transcript_errors}, the hypothesis $X$ has 3 errors against the reference, $Y$ and $Z$, causing difficulties in aligning them:

\begin{enumerate}
\setlength\itemsep{0em}

\item A spelling and word recognition error; `\textit{going}' is recognized as  `\textit{gonna}' in the hypothesis.
\item A missing word; `\textit{uh}' is not recognized.
\item Overlapped utterances; B's utterance is spoken while A utters `\textit{Amsterdam}', which are merged into one utterance for A'.
\end{enumerate}

\noindent The first two types are ASR errors that can be handled by most pairwise alignment methods such as the Needleman-Wunsch (NW) algorithm \cite{NEEDLEMAN1970}.
However, the third type is an SD error involving multiple sequences, occurring when utterances by distinct speakers overlap in time.
Figure~\ref{fig:pairwise_result} illustrates how the NW algorithm treats them as insertion and deletion errors, leading to incomplete alignment of those tokens:

\begin{figure}[htbp!]
    \centering
    \includegraphics[width=\columnwidth]{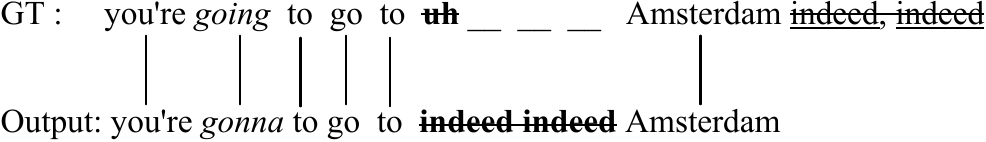}
    \caption{The result by the NW algorithm.}
    \label{fig:pairwise_result}
\vspace{-0.5em}
\end{figure}

\begin{figure}[htbp!]
    \centering
    \includegraphics[width=\columnwidth]{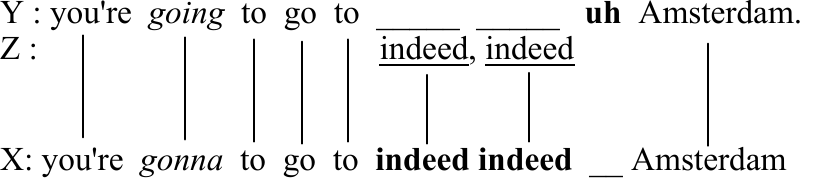}
    \caption{The result by our multi-sequence alignment algorithm for the above example.}
    \label{fig:MSA_result}
\vspace{-1em}
\end{figure}

\noindent To overcome this challenge, a new multi-sequence alignment algorithm is designed by increasing the dimension of dynamic programming, allowing us to process utterances from all sequences in parallel (Figure~\ref{fig:MSA_result}).
Our algorithm shares a similar idea with the one by \citet{multi-dimension} for expanding the reference into multiple sequences based on speakers and applying multi-dimensional dynamic programming to solve the alignment problem.
While their solution is based on the Levenshtein distance with the aid of a directed acyclic graph, however, our algorithm uses the NW algorithm for efficiency.
Furthermore, we use a different scoring criteria consisting of fully match, partially match, mismatch, and gap that we find more effective, whereas they use match, insertion, deletion, and substitution.

\subsection{Algorithm: Scoring Matrix}
\label{ssec:algorithm-scoring-matrix}

Our multi-sequence alignment algorithm extends the NW algorithm to handle multiple dimensions by enhancing the scoring matrix and backtracking strategy.
Let $X = [x_1, .., x_\ell]$ be a sequence created by listing all tokens in the hypothesis transcript regardless of segmentation.
Let $Y_j = [y_{j1}, .., y_{jm}]$ be a sequence created by listing tokens of Speaker $Y_j$,\LN the $j$'th speaker, in the reference transcript.
Given $E = [X, Y_1, .., Y_n]$, the algorithm first populates the scoring matrix $F$, a multidimensional matrix whose dimensions are determined by the input sequence lengths, where all cells are initialized to~$0$:

\begin{algorithm}[!htbp]
\SetAlgoLined
\SetKwInOut{Input}{Input}
\SetKwInOut{Output}{Output}
\Input{$E = \{X, Y_1, \ldots, Y_n\}$}
\Output{The scoring matrix $F$}
Create $F \in \mathbb{R}^{(|X|+1) \times (|Y_1|+1) \times \cdots \times (|Y_n|+1)}$\;
$C \gets [\gamma \subset \{0, 1, \ldots, n\}] \setminus \varnothing$\;
\ForEach{$\gamma \in C$}{
    \ForEach{$\psi \in \textit{index\_perm}(\gamma, E)$}{
        $F_{\psi} \gets \textit{score}(\psi, E, F)$\;
    }
}
\Return{F}\;
\caption{Scoring Matrix Population}
\label{alg:MSAwithPermutations}
\end{algorithm}

\noindent Algorithm~\ref{alg:MSAwithPermutations} illustrates how the scoring matrix is populated, which is generalizable to any number of sequences.
Once the scoring matrix is created (\texttt{L1}), it generates a list comprising all combinations of $\{0, .., n\}$ expect for the empty set (\texttt{L2}).
For the popular case of 2-speaker dialogues where $n = 2$, $C$ is generated as follows:
$$
[\{0\}, \{1\}, \{2\}, \{0, 1\}, \{0, 2\}, \{1, 2\}, \{0, 1, 2\}]
$$
Note that the order of subsets in $C$ matters because the indices produced by earlier combinations must be processed before the later ones.
The numbers in a combination represent the input sequences, where $0$ represents $X$ and $i$ represents $Y_i$ (${\forall i>0}$).
Each combination $\gamma$ and $E$ are passed to the \textit{index\_perm} function that returns a list of index tuples (\texttt{L3-4}). 
The tuples are generated with indices for the corresponding sequences while the indices for the other sequences remain at $0$.
Table~\ref{tab:index-perm} describes the lists of\LN index tuples for the above combinations.

\begin{table}[htbp!]
\centering\resizebox{\columnwidth}{!}{
\begin{tabular}{c|c|c} 
\toprule
$\bm{\gamma}$ & $\bm{\textit{index\_perm}(\gamma, E)}$ & \bf Size \\
\midrule
$\{0\}$       & $[(1, 0, 0), .., (|X|, 0, 0)]$         & $|X|$ \\
$\{1\}$       & $[(0, 1, 0), .., (0, |Y_1|, 0)]$       & $|Y_1|$ \\
$\{2\}$       & $[(0, 0, 1), .., (0, 0, |Y_2|)]$       & $|Y_2|$ \\
$\{0, 1\}$    & $[(1, 1, 0), .., (|X|, |Y_1|, 0)]$     & $|X| \cdot |Y_1|$ \\
$\{0, 2\}$    & $[(1, 0, 1), .., (|X|, 0, |Y_2|)]$     & $|X| \cdot |Y_2|$ \\
$\{1, 2\}$    & $[(0, 1, 1), .., (0, |Y_1|, |Y_2|)]$   & $|Y_1| \cdot |Y_2|$ \\
$\{1, 2, 3\}$ & $[(1, 1, 1), .., (|X|, |Y_1|, |Y_2|)]$ & $|X| \cdot |Y_1| \cdot |Y_2|$ \\ 
\bottomrule
\end{tabular}}
\caption{Generating permutations of index tuples.}
\label{tab:index-perm}
\vspace{-1em}
\end{table}

\noindent For each index tuple $\psi = (i, j, .., k)$ where $i$ indicates $x_i \in X$, $j$ indicates $y_{1j} \in Y_1$, and $k$ indicates $y_{nk} \in Y_n$, the \textit{score} function considers the scores of all cells straightly prior to $x_i$ such as:
$$\{(i-1, j, .., k), (i, j-1, .., k), \ldots, (i, j, .., k-1)\}$$
or diagonally prior to $x_i$ such as:
$$\{(i-1, j-1, .., k), \ldots, (i-1, j, .., k-1)\}$$
and measures the score of $F_{\psi}$ as follows (\texttt{L5}):
\begin{equation}
\label{eq:3d-scoring-function}
\small
   \begin{array}{ccl}
   F_{i, j, .., k} & \gets & \max(\mathcal{G}(E, F, (i, j, .., k))) \\[0.5em]
   \mathcal{G}(E, F, \psi) &\gets & \begin{cases}
      \begin{split}
         F_{i-1,   j, .., k}   &+ \textit{match}(x_i) \\
         F_{  i, j-1, .., k}   &+ \textit{match}(y_{1j}) \\
         &\:\:\vdots \\
         F_{  i,   j, .., k-1} &+ \textit{match}(y_{nk})\\
         F_{i-1, j-1, .., k}   &+ \textit{match}(x_i, y_{1j}) \\
         &\:\:\vdots \\
         F_{i-1,   j, .., k-1} &+ \textit{match}(x_i, y_{nk}) \\
      \end{split}
      \end{cases}
   \end{array}
\end{equation}
The \textit{match} function returns $-1$ if only one token from $E$ is passed, indicating that it can be matched only with gaps that are artificially inserted to handle tokens not finding any match with ones in the other sequences.
If two tokens are passed, it measures the Levenshtein Distance ($LD$) between them and returns the value as follows:\footnote{For our experiments, $d=1$ is used.}
\[\small
\textit{match}(x,y) \gets
  \begin{cases}
    2  & \text{if $LD(x,y) = 0$ (fully match)}\\
    1  & \text{if $LD(x,y) \leq d$ (partial match)}\\
    -1 & \text{if $LD(x,y) > d$ (mismatch)}\\
  \end{cases}
\]
Note that when two tokens are passed to the \textit{match} function, one of them must be $x_i$ so that it always\LN compares a token in $X$ (hypothesis) with another token in $Y_*$ (reference), but never compares two tokens in $Y_*$ (e.g., $\textit{match}(y_{1j}, y_{nk})$) that are both from the reference.
Moreover, the algorithm does not allow $x_i$ to match with multiple tokens in $Y_*$ (e.g., \textit{match}$(x_i, y_{1j}, y_{nk})$).
Although it is possible for two speakers to say the exact same token at the same time, it is rare and considered an exception.

\subsection{Algorithm: Backtracking}
\label{ssec:algorithm-backtracking}

Algorithm~\ref{alg:backtracking} outlines our backtracking strategy that takes the list of input sequences $E$ and the scoring matrix $F$ in Section~\ref{ssec:algorithm-scoring-matrix}, and returns the alignment matrix $A$.
It creates $A$, where the $0$'th and $i$'th rows will be filled with tokens in $X$ and $Y_i$ respectively or gap tokens (\texttt{L1}).\footnote{The value of $\rho$ cannot be determined at this stage because the number of gap tokens needed for the alignment is unknown until the backtracking process is completed.}
Thus, the number of columns $\rho = \max(|X|+g_x, |Y_i|+g_i : \forall i)$, where $g_x$ and $g_i$\LN are the numbers of gap tokens inserted to find the best alignment for $X$ and $Y_i$, respectively.

\begin{algorithm}[!htbp]
\SetAlgoLined
\SetKwInOut{Input}{Input}
\SetKwInOut{Output}{Output}
\Input{$E = \{X, Y_1, \ldots, Y_n\}$,\\ the scoring matrix $F$.}
\Output{The alignment matrix $A$}
Create $A \in \mathbb{R}^{|E| \times \rho}$\;
$\psi \gets (|X|, |Y_1|, \ldots, |Y_n|)$\;
\While{$\psi \neq (0, 0, \ldots, 0)$}{
    $(\psi', \alpha) \gets \textit{argmax}(\mathcal{G}(E, F, \psi))$\;
    Append $\alpha$ to $A$ accordingly\;
    $\psi \gets \psi'$\;
}
\Return{A}\;
\caption{Backtracking Strategy}
\label{alg:backtracking}
\end{algorithm}

\begin{figure}[htbp!]
    \centering
    \includegraphics[width=0.9\columnwidth]{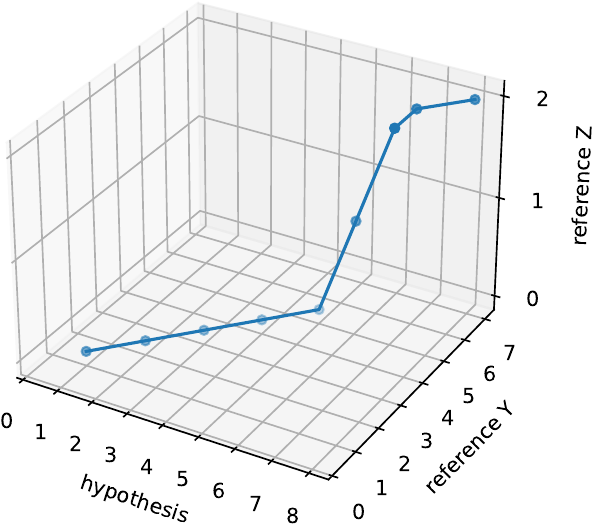}
    \caption{The backtracking example using Algorithm~\ref{alg:backtracking} and the scoring matrix in Table~\ref{tab:scoring-matrix-3d} (Appendix~\ref{ssec:msa-demonstration}).}
    \label{fig:backtracking}
\vspace{-1em}
\end{figure}

\noindent The backtracking process starts from the last cell indexed by $\psi$ (\texttt{L2}).
It then finds a cell (\texttt{L4}) using the \textit{argmax} function, which returns the index tuple $\psi'$\LN and the token list $\alpha$ that maximize the alignment score ($|\alpha| = |E|$).
The $0|i$'th item in $\alpha$ is either the currently visited token in $X|Y_i$ or a gap token `$-$'.
For example, among the conditions in $\mathcal{G}(E, F, \psi)$ (Eqn.~\ref{eq:3d-scoring-function}), suppose that $F_{i-1, j-1, .., k} + \textit{match}(x_i, y_{1j})$ provides the highest score. In this case, it returns $\psi' = (i-1, j-1, .., k)$ and $\alpha = [x_i, y_{1j}, ..., -]$.
The tokens in $\alpha$ are appended to the corresponding sequences (\texttt{L5}).
For the above example, tokens in $\alpha$ are appended to $A$ as follows:
$$
\begin{array}{ccccl}
A_0 & \gets & A_0 & \oplus & [x_i]\\
A_1 & \gets & A_1 & \oplus & [y_{1j}]\\
    & \vdots & & & \\
A_2 & \gets & A_n & \oplus & [-]
\end{array}
$$
Finally, it moves to the next cell indexed by $\psi'$ (\texttt{L6}).
This process continues until the algorithm reaches the first cell (\texttt{L3}).
Figure~\ref{fig:backtracking} shows the backtracking performed by Algorithm~\ref{alg:backtracking} using the scoring matrix produced by Algorithm~\ref{alg:MSAwithPermutations} (Table~\ref{tab:scoring-matrix-3d}; Appendix~\ref{ssec:msa-demonstration}) for the working example.
The resulting alignment matrix of this example is presented in Table~\ref{tab:alignment-matrix} (\ref{ssec:msa-demonstration}).

\subsection{Optimization}
\label{ssec:optimization}

To conserve memory when aligning sequences with a large number of speakers/tokens, a segmentation method is implemented.
This involves segmenting the dialogue into smaller chunks based on detecting short absolutely aligned segments as barriers, with the length of each segment set to a given minimum. 
The segmentation is performed at the mid-point of each barrier, and each segment is aligned separately.
This approach limits the maximum memory usage.

It is worth mentioning that the number of cells in the scoring matrix that the original NW algorithm compares is the sum of all combinations ($n' = |E|$), $\sum_{i=1}^{n'} C(n', i)=2^{n'}-1$.
However, our algorithm matches the hypothesis tokens with only necessary tokens in the reference, as cells that do not involve the hypothesis token or involve more than two non-gap tokens are ignored (Section~\ref{ssec:algorithm-scoring-matrix}).
This reduces the number of cells for each comparison to $2\cdot n' - 1$, greatly reducing the decoding time consumed.

\section{Text-based SD Evaluation}
\label{sec:text-based-sd-evaluation}

Section~\ref{sec:related-work} addresses the limitations and challenges of existing metrics for evaluating SD. 
To overcome these issues, two metrics are proposed, Text-based Diarization Error Rate (Section~\ref{ssec:tder}) and diarization F1 (Section~\ref{ssec:d-f1}), which are made possible by the token alignment achieved in Section~\ref{sec:align4d}.

\subsection{Text-based Diarization Error Rate}
\label{ssec:tder}

The original \texttt{DER} quantifies the amount of time, in which an audio segment is incorrectly assigned to a speaker (Section~\ref{ssec:der}).
For text-based evaluation, the duration of the audio can be directly translated into the sequence length, i.e., the number of tokens in the sequence. 
This allows us to estimate several types of SD errors by examining tokens aligned to incorrect speakers or gap tokens.
With these adaptations, Text-based Diarization Error Rate (\texttt{TDER}) can be formulated as follows:
\begin{equation}
\label{eq:TDER-full}
\small
    \texttt{TDER} =
    \frac{\sum_{\forall u \in U} len(u) \cdot (\max(N_{r}(u), N_{h}(u)) - N_{c}(u))}{\sum_{\forall u \in U} len(u) \cdot N_{r}(u)}
\end{equation}

\noindent $U = [u_1, .., u_q]$ is the reference transcript where $u_i$ is the $i$'th utterance in $U$.
$len(u)$ is the number of tokens in $u$.
$N_{r}(u)$ and $N_{h}(u)$ are the numbers of speakers in $u$ given the reference and hypothesis transcripts, respectively.
$N_{c}(u)$ is the number of correctly identified speakers in $u$.
Unlike an audio segment that can involve multiple speakers, a text utterance in the reference transcript is always spoken by one speaker, so $N_{r}(u) = 1$.
Hence, \texttt{TDER} can be rewritten as follows:
\begin{equation}
\label{eq:TDER}
\small
    \texttt{TDER} = \frac{\sum_{\forall u \in U} len(u)\cdot (\max(1, N_{h}(u)) - N_{c}(u))}{\sum_{\forall u \in U} len(u)}
\end{equation}

\noindent \texttt{TDER} captures different types of SD errors:

\begin{itemize}
\item \textit{Speaker errors} ($E_{se}$) are detected in the scenario when $N_{h}(u) = 1$ and $N_{c}(u) = 0$.
\item When $N_{h}(u) > 1$, it indicates \textit{false alarm} errors\LN ($E_{fa}$). False alarm errors in text-based SD mostly occur when the system identifies certain parts of an utterance not correctly as spoken by different speakers.
These errors can occur for non-speech segments if the system includes them in the transcript (e.g., ``Hello, (\textit{pause}) how are you?'').
\item $N_{h}(u) = 0$ implies \textit{missed speech} errors ($E_{ms}$) in which case, the system misses to transcribe those segments of the audio.
\item Like \texttt{DER}, \textit{overlapping speech} errors result in $N_{h}(u) = 0$; thus, they are included in $E_{ms}$. Note that in an audio segment containing overlapping speeches, the corresponding text transcript may include multiple utterances, while ASR systems usually transcribe only one of them, so the untranscribed utterances are considered ``missed''.
\end{itemize}

\noindent \texttt{TDER} can handle any number of sequences in the reference transcript, as well as the situation when the hypothesis contains a different number of tokens from the reference.
Compared to the existing evaluation metrics such as \texttt{WDER} (Section~\ref{ssec:wder}) or \texttt{WER} (Section~\ref{ssec:wer}), \texttt{TDER} assesses a greater variety of error types, making it easy to perform a comprehensive analysis in text-based speaker diarization.

\subsection{Diarization F1}
\label{ssec:d-f1}

While \texttt{TDER} is a comprehensive evaluation metric, it only considers utterances in the reference and ignores tokens in the hypothesis that are not aligned to any tokens in the reference.
Hence, \texttt{TDER} does not penalize the inclusion of additional tokens in the hypothesis that do not correspond to the audio.
To address this limitation, we propose Diarization F1 (\texttt{DF1}), which performs token-level analysis by measuring precision and recall, i.e., how many tokens in the hypothesis and reference are correctly identified with speakers, respectively:
\begin{equation}
\label{eq:df1}
\small
\begin{split}
    \text{Precision} &= \frac{|speaker\_match(T_{r}, T_{h})|}{|T_{h}|}\\[0.3em]
    \text{Recall}    &= \frac{|speaker\_match(T_{r}, T_{h})|}{|T_{r}|}
\end{split}
\end{equation}

\noindent $T_r$ and $T_h$ are sequences of tokens in the reference and hypothesis transcripts, respectively.
The function $\textit{speaker\_match}(T_r, T_h)$ returns a sequence of tokens in $T_r$, say $T'_r$, such that each token $t_r \in T'_r$ is aligned with some token $t_h \in T_h$, and the speaker of $t_r$ by the reference is the same as the speaker identified for $t_h$ by the hypothesis.

\section{Experiments}
\label{sec:experiments}

\subsection{Automatic Transcribers and Corpus}
\label{ssec:transcribers-corpus}

While there are many automatic transcribers, most of them do not perform SD \cite{NEURIPS2020_92d1e1eb,whisper} so that only limited off-the-shelf options are available.
For our experiments, we\LN use two transcribers, Amazon Transcribe and Rev AI, which are publicly available, can perform both ASR and SD, and offer a free tier of usage,\footnote{\url{https://aws.amazon.com/transcribe}\\\url{https://www.rev.ai}} making them accessible options for our study.

We use the CABank English CallHome Corpus \cite{callhome}, comprising 120 unscripted, informal telephone conversations between native English speakers that cover various topics.
Their transcripts follow the CHAT (Codes for the Human\LN Analysis of Transcripts) format, capturing several aspects of spoken language such as speaker turns, pauses, overlapping speech, and non-verbal cues.

For evaluation, we manually select 10 conversations from this corpus based on their audio quality. 
Each conversation lasts approximately 30 minutes, but the reference transcript only covers the first 10 minutes.
Thus, each audio is cut into a 10-minute segment and transcribed by the above two systems.

\subsection{Speech Recognition Evaluation}
\label{ssec:asr-evaluation}

Building on the previous work \cite{WER}, we assess the quality of transcripts produced by the two transcribers in Section~\ref{ssec:transcribers-corpus}, and analyze the following four types of ASR errors:
\begin{itemize}
\setlength\itemsep{0em}
\item \textbf{Missing Token}: It occurs when a token present in the reference is not detected by the system, resulting in a missing token in the hypothesis.
\item \textbf{Extra Token}: It occurs when the system inserts an extra token not present in the reference, resulting in an extra token in the hypothesis.
\item \textbf{Substitution}: It occurs when the system replaces a token in the reference with a different token in the hypothesis.
\item \textbf{Overlapping}: It occurs when two or more speakers talk at the same time so that the system cannot accurately transcribe all speakers.
\end{itemize}

\noindent Table~\ref{table:transcriber-errors} summarizes the error distributions for Amazon Transcribe (AT) and Rev AI (RA). 
AT exhibits a higher rate of \textit{missing tokens}, suggesting that it skips audio segments that are unclear or difficult to recognize, which leads to omit the corresponding tokens. 
In contrast, RA has higher rates in the other three error types, implying that it tends to preserve most of the information. 
This is also reflected in the \textit{overlapping} where AT transcribes no overlapping tokens while RA does, making it more challenging to accurately transcribe and susceptible to errors.

\begin{table}[htbp!]
\centering\small
\begin{tabular}{c|cccc|c}
    \toprule
    \bf Transcriber & \bf MT & \bf ET & \bf ST & \bf OL & $\bm{\sum}$ \\
    \midrule
    Amazon (AT) &     6.8 & \bf 1.5 & \bf 2.8 & \bf 0.0 & \bf 11.1 \\
    Rev AI (RA) & \bf 5.1 &     2.7 &     3.1 &     0.5 &     11.4 \\
    \bottomrule
\end{tabular}
\caption{Average percentages of the four types of errors over all tokens. MT: missing tokens, ET: extra tokens, ST: substitutions, OL: overlapping.}
\label{table:transcriber-errors}
\vspace{-1em}
\end{table}

\begin{figure*}[htbp!]
\centering
\includegraphics[width=\textwidth]{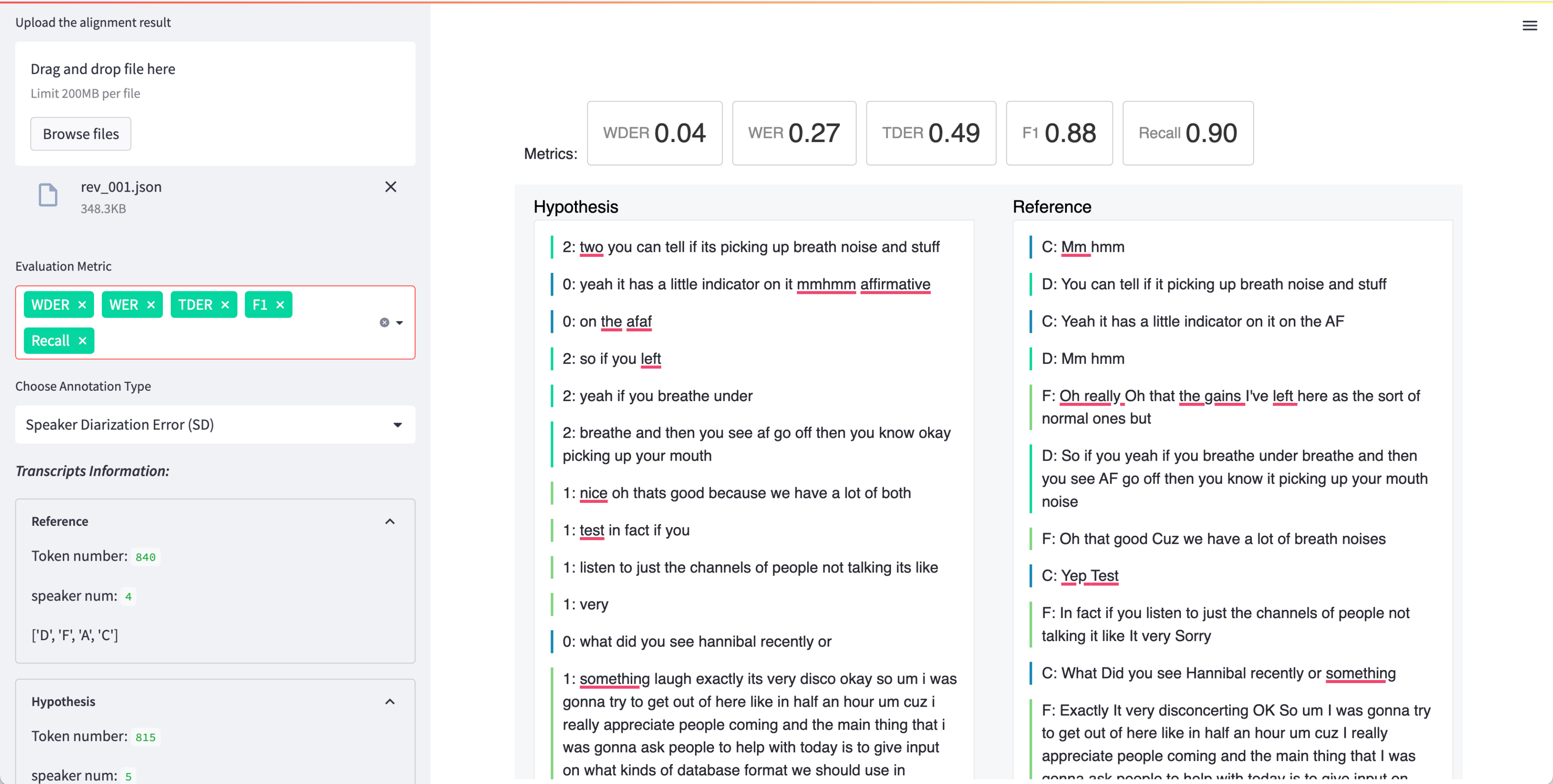}
\caption{A screenshot of our visualization tool, TranscribeView. Given the reference and hypothesis transcripts, it simultaneously displays all sequences and allows us to view token alignments as well as selected ARR/SD errors.}
\label{fig:Interface}
\vspace{-1em}
\end{figure*}

\subsection{Token Alignment Evaluation}
\label{ssec:token-alignment-evaluation}

To evaluate the robustness of our multiple sequence alignment algorithm (MSA; Section~\ref{sec:align4d}), we employ hypothesis transcripts by AT and RA and measure the proportions of correctly aligned tokens in the reference transcripts.
The final accuracy is obtained by averaging the results from all 10 transcripts. 
The MSA performance is compared with the character-level (the original NW) algorithm and also a token-level alignment algorithm without multi-sequence support, which is MSA restricted to utilize only a 2-dimensional scoring matrix and linearize multiple sequences in the reference as in Figure~\ref{fig:pairwise_result}.

\noindent Table~\ref{table:alignmentAcc} demonstrates a significant improvement in the performance of MSA compared to the other two algorithms, ensuring its robustness for adaptation in text-based SD evaluation (Section~\ref{sec:text-based-sd-evaluation}).

\begin{table}[htbp!]
\centering\small
\begin{tabular}{c|c}
\toprule
\bf Alagorithm & \bf Accuracy \\
\midrule
Character-level (original NW) & 0.92 \\
Token-level w/o Multi-Seq.\ Support & 0.93 \\
Token-level with Multi-Seq.\ Support & \bf 0.99 \\
\bottomrule
\end{tabular}
\caption{Average accuracies achieved by three types of alignment algorithms on AT and RA transcripts.}
\label{table:alignmentAcc}
\vspace{-1em}
\end{table}

\subsection{Speaker Diarization Evaluation}

To evaluate text-based SD, we use the Hungarian algorithm, an efficient method for finding the optimal\LN assignment in a cost matrix \cite{kuhn1955hungarian}. 
In our case, the cost matrix reflects the errors in assigning reference speakers to hypothesis speakers.
It then determines the optimal assignment by minimizing the total cost based on the cost matrix.

\begin{table}[htbp!]
\centering\small
\begin{tabular}{c|ccc|cc}
    \toprule
\bf Transcriber & \textbf{\texttt{DER}} & \textbf{\texttt{WDER}} & \textbf{\texttt{WER}}  & \textbf{\texttt{TDER}} & \textbf{\texttt{DF1}}  \\ 
\midrule
    Amazon (AT) & \bf 0.24 & \bf 0.15 &     0.34 &     0.53 &     0.79 \\ 
    Rev AI (RA) &     0.26 &     0.20 & \bf 0.29 & \bf 0.50 & \bf 0.84 \\ 
\bottomrule
\end{tabular}
\caption{Comparing the traditional metrics (\texttt{DER}, \texttt{WDER}, \texttt{WER}) with our new evaluation metrics (\texttt{TDER}, \texttt{DF1}).}
\label{table:result}
\vspace{-1em}
\end{table}

\noindent Table~\ref{table:result} shows the average scores measured by our new metrics, \texttt{TDER} (\textsection\ref{ssec:tder}) and \texttt{DF1} (\textsection\ref{ssec:d-f1}), as well as the traditional metrics, \texttt{DER} (\textsection\ref{ssec:der}), \texttt{WDER} (\textsection\ref{ssec:wder}), and \texttt{WER} (\textsection\ref{ssec:wer}) on the hypothesis transcripts for the 10 conversations.
AT performs better than RA in SD as shown by lower \texttt{DER} and \texttt{WDER}, indicating that it accurately segments different speakers at the audio level. 
Conversely, RA performs better than AT in ASR, as evidenced by the lower \texttt{WER}.

The (Precision, Recall) scores for \texttt{DF1} are (0.87, 0.73) and (0.88, 0.81) for AT and RA, respectively.
Notably, \texttt{WDER} only considers aligned tokens (as explained in Section~\ref{ssec:wder}).
Moreover, Section~\ref{ssec:asr-evaluation} observes AT's tendency to omit tokens during ASR, which can cause a lower recall score and a skewed \texttt{WDER} result.
Due to this, AT has a lower \texttt{DF1} score than RA contributed by a low recall score of 0.73.

While traditional SD metrics such as \texttt{DER} and \texttt{WDER} indicate an advantage for AT over RA, our new metrics, \texttt{TDER} and \texttt{TF1}, reveal RA's superiority in text-based SD performance. 
This highlights the strength of our new metrics, as they evaluate SD at the utterance-level, providing a more accurate reflection of the diarization quality than the traditional metrics, which are based on uniformly-split audio segments or word-level analysis.\footnote{More details of this analysis are provided in Appendix~\ref{ssec:transcribeview-demonstration}.}

\section{Applications}
\label{sec:applications}

We offer two tools to facilitate the adaptation of our work: \textbf{align4d}, an efficient MSA tool with an user-friendly API (Section~\ref{ssec:align4d}) and \textbf{TranscribeView}, a visualization interface to analyze ASR/SD errors through multiple evaluatioin metrics (Section~\ref{ssec:transcribeview}).
These tools enable us to thoroughly analyze those errors and create higher-quality transcript data.

\begin{figure*}[htbp!]
\centering
\includegraphics[width=\textwidth]{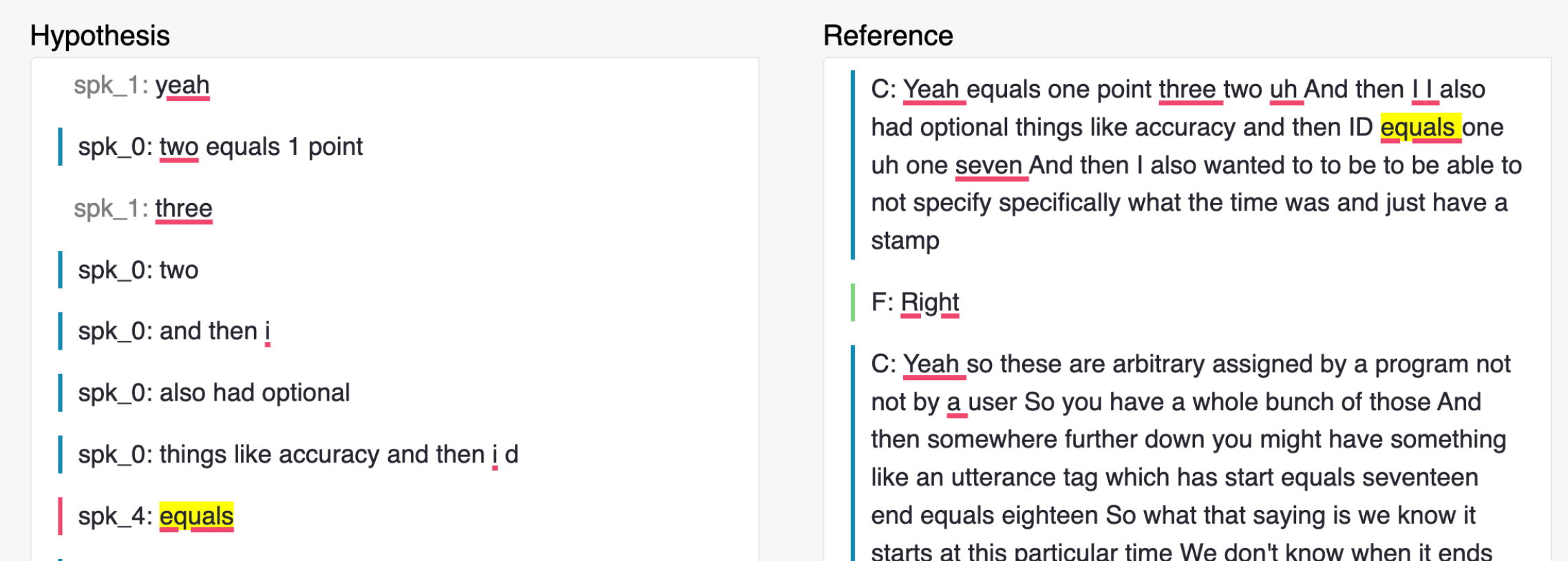}
\caption{A screenshot of the alignment area in TranscribeView. Each vertical colored bar represents the alignment between speakers (e.g., $spk_0$ is aligned to $C$), while greyed-out speaker labels indicate unmapped speakers (e.g., $spk_1$ is not mapped to any speakers in the reference). Users can hover over tokens to view their corresponding aligned token (highlighted in yellow). Diarization errors are indicated by red underlines.}
\label{fig:amazon-case}
\vspace{-1em}
\end{figure*}

\subsection{Align4d}
\label{ssec:align4d}

Our MSA algorithm can be computationally intensive due to its creation of a high-dimensional matrix and an exhaustive search to find the global optimum through dynamic programming.
To improve its efficiency, we have implemented the algorithm in C++ and compiled it as a CPython extension, which can be imported as a Python package.
To enhance its adaptability, the C++ dependencies are restricted to\LN the C++20 standard template library.

Our Python API takes reference and hypothesis sequences in JSON as input.
It also allows users to strip punctuation, and parameterize the Levenshtein Distance for matching (\textsection\ref{ssec:algorithm-scoring-matrix}) and the segmentation length for optimization (\textsection\ref{ssec:optimization}).
Finally, it returns the alignment matrix in JSON (\textsection\ref{ssec:algorithm-backtracking}).
Based on our\LN testing, {align4d} can comfortably handle 200,000 tokens involving 5 speakers using an average laptop.
It is publicly available as an open source API:\\ \url{https://github.com/anonymous/align4d}.

\subsection{TranscribeView}
\label{ssec:transcribeview}

Figure~\ref{fig:Interface} shows the graphical user interface of TranscribeView, which offers a comprehensive analysis of ASR and SD. 
This tool uses align4d (Section~\ref{ssec:align4d}) to align tokens from the reference and hypothesis transcripts and presents them side-by-side for easy comparison. 
It also provides statistical information about the transcripts, such as the number of tokens and speakers (Figure~\ref{fig:sidebars} in Appendix~\ref{ssec:transcribeview-demonstration}). 
Users can select evaluation metrics from the following: \texttt{WDER} (\textsection\ref{ssec:wder}), \texttt{WER} (\textsection\ref{ssec:wer}), \texttt{TDER} (\textsection\ref{ssec:tder}), and \texttt{DF1} (\textsection\ref{ssec:d-f1}).
The evaluation scores are displayed at the top of the alignment area, in which every utterance is marked by a virtual colored bar implying the corresponding speaker.
Users can also hover over tokens to see the\LN corresponding aligned tokens.
Moreover, SD errors are indicated by red underlines.

TranscribeView is a web-based application built using the Streamlit framework with custom HTML elements such that it can be accessed using any web\LN browser.
It is publicly available as an open-source\LN software: \url{https://github.com/anonymous/TranscribeView}.

\section{Conclusion}
\label{sec:conclusion}

This paper presents a novel approach to evaluating text-based speaker diarization by introducing two metrics, Text-based Diarization Error Rate (\texttt{TDER}) and Diarization F1 (\texttt{DF1}), along with an enhanced algorithm for aligning transcripts with multiple sequences.
Our multiple sequence alignment (MSA) algorithm enables accurate token-to-token mapping between reference and hypothesis transcripts.
Our web-based tool, TranscribeView, provides a comprehensive platform that allows researchers to visualize and evaluate errors in speech recognition as well as speaker diarization.

While this work provides valuable contributions, it also recognizes a few limitations.
The robustness of our alignment algorithm and the effectiveness of our proposed evaluation metrics can be further verified by annotating more transcripts, which is labor-intensive.
The increased computational complexity from the enhanced MSA algorithm may also limit\LN  its applicability.
Future work aims to improve the efficiency and effectiveness of TranscribeView and align4d to handle a wider range of research.


\bibliography{custom}

\cleardoublepage\appendix
\section{Appendix}
\label{sec:appendix}

\subsection{TranscribeView Demonstration}
\label{ssec:transcribeview-demonstration}

\begin{figure}[htbp!]
  \centering
  \begin{subfigure}[b]{0.45\columnwidth}
    \includegraphics[width=\columnwidth]{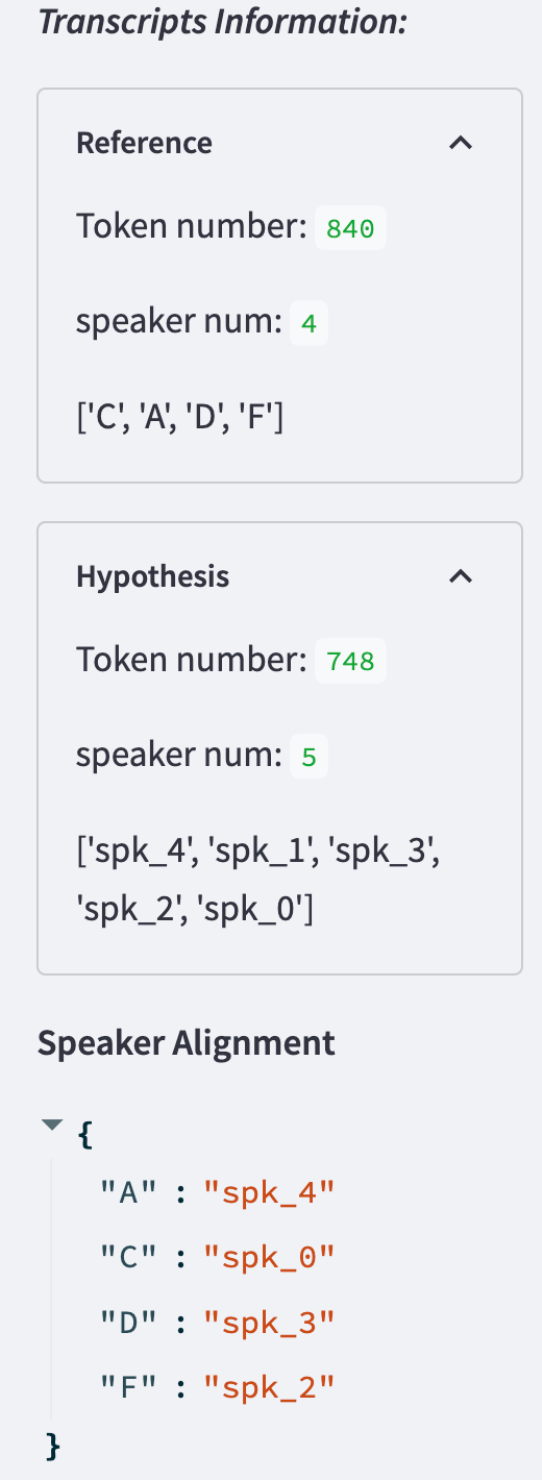}
    \caption{Amazon Transcribe}
    \label{fig:plot2}
  \end{subfigure}
  \hfill
  \begin{subfigure}[b]{0.45\columnwidth}
    \includegraphics[width=\columnwidth]{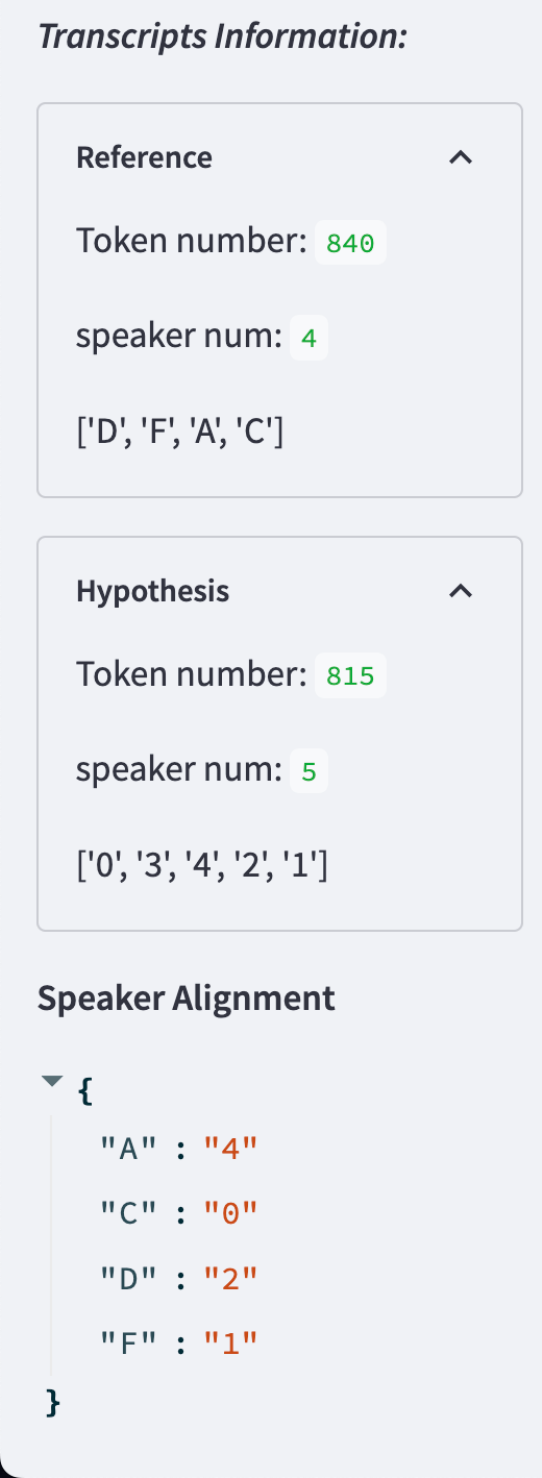}
    \caption{Rev AI}
    \label{fig:plot1}
  \end{subfigure}
  \caption{Amazon Transcribe and Rev AI's transcript information after uploading the input JSON file.}
  \label{fig:sidebars}
\end{figure}

In TranscribeView (Section~\ref{ssec:transcribeview}), the transcript information is summarized at the bottom of the sidebar once the input JSON file, containing reference and hypothesis transcripts, is uploaded.
In Figure~\ref{fig:sidebars}, the reference transcript consists of 840 tokens with 4 speakers.
The outputs of Amazon Transcribe (AT) and Rev AI (RA) contain 748 and 815, respectively.
Both tools produce fewer numbers of tokens compared to the reference, although AT drops significantly more, approximately $11\%$ of the reference.
Moreover, both tools identify 5 speakers, although there are only 4 speakers in the original audio.
As a result, one of the speakers in each hypothesis transcript cannot be aligned to any reference speaker (e.g., `$spk_1$' for AT and `$3$' for RA).

\begin{figure}[htbp!]
  \centering
  \begin{subfigure}[b]{\columnwidth}
    \includegraphics[width=\columnwidth]{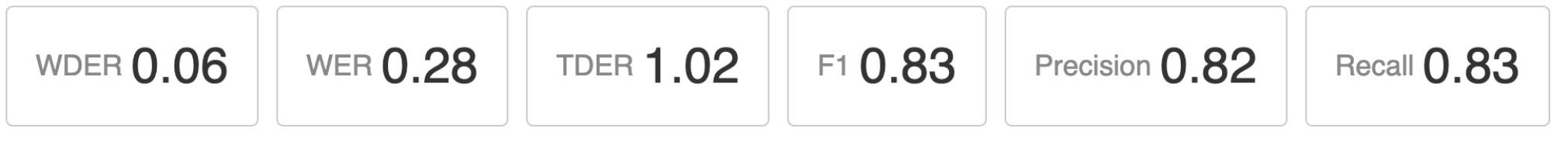}
    \caption{Amazon Transcribe}
    \label{fig:plot2}
  \end{subfigure}
  \hfill
  \begin{subfigure}[b]{\columnwidth}
    \includegraphics[width=\columnwidth]{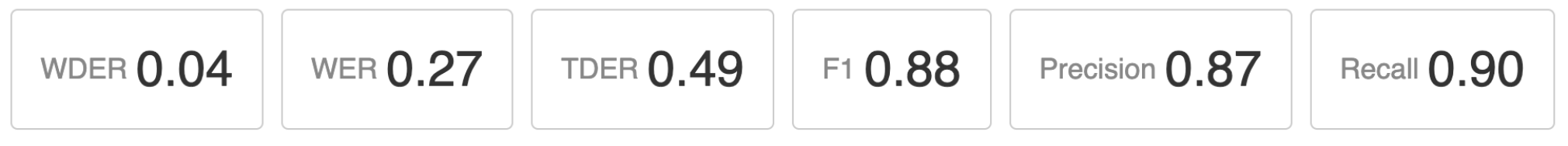}
    \caption{Rev AI}
    \label{fig:plot1}
  \end{subfigure}
  \caption{Screenshots of the evaluation metric area.}
  \label{fig:metric-screenshot}
\end{figure}

\noindent The selected evaluation metrics are displayed at the top of the visualization area.
Figure~\ref{fig:metric-screenshot} shows the selected evaluation metrics for both transcribers.
When examining \texttt{WDER} and \texttt{WER}, the performance of the transcribers appears similar for both ASR and SD.
However, \texttt{TDER} and \texttt{DF1} reveal a significant difference in SD quality.
The higher number of dropped tokens in AT's output contributes to a noticeable discrepancy in recall.
This error is disregarded in \texttt{WDER}, which only considers errors in aligned tokens.
Hence, based on our new metrics, RA's transcript exhibits higher quality, particularly for SD, which agrees with our manual evaluation on\LN these outputs.

\subsection{MSA Demonstration}
\label{ssec:msa-demonstration}

Table~\ref{tab:scoring-matrix-3d} illustrates the scoring matrix $F \in \mathbb{R}^{9 \times 8 \times 3}$ generated by Algorithm~\ref{alg:MSAwithPermutations} for the working example\LN in Section~\ref{sec:align4d}, where \texttt{SOL} is a meta-token prepended to every sequence and represents the start-of-line.
All punctuation marks are stripped from the input before processing.
Given the scoring matrix, Algorithm~\ref{alg:backtracking} takes the following backtracking steps to create the alignment matrix in Table~\ref{tab:alignment-matrix}:

\begin{enumerate}
\item It begins at the last cell, $(x_8, y_7, z_2)$.
\item It finds \textbf{10} by matching $(x_8, y_7)$, and moves to $(x_7, y_6, z_2)$.
\item It finds \textbf{11} by matching $y_6$ to gaps, and moves to $(x_7, y_5, z_2)$.
\item It finds \textbf{9} by matching $(x_7, z_2)$, and moves to $(x_6, y_5, z_1)$.
\item It finds \textbf{7} by matching $(x_6, z_1)$, and moves to $(x_5, y_5, z_0)$.
\item It finds \textbf{5} by matching $(x_5, y_5)$, and moves to $(x_4, y_4, z_0)$.
\item It finds \textbf{3} by matching $(x_4, y_4)$, and moves to $(x_3, y_3, z_0)$.
\item It finds \textbf{1} by matching $(x_3, y_3)$, and moves to $(x_2, y_2, z_0)$.
\item It finds \textbf{2} by matching $(x_2, y_2)$, and moves to $(x_1, y_1, z_0)$.
\item It finds \textbf{0} by matching $(x_1, y_1)$, and moves to $(x_0, y_0, z_0)$.
\item The first cell, $(x_0, y_0, z_0)$, is reached; therefore, it terminates.
\end{enumerate}

\begin{table*}[htbp!]
\centering

\begin{subtable}{\textwidth}
\centering\small{ 
\begin{tabular}{cc|>{\centering\arraybackslash}p{0.8cm}>{\centering\arraybackslash}p{0.8cm}>{\centering\arraybackslash}p{0.8cm}>{\centering\arraybackslash}p{0.8cm}>{\centering\arraybackslash}p{0.8cm}>{\centering\arraybackslash}p{0.8cm}>{\centering\arraybackslash}p{0.8cm}>{\centering\arraybackslash}p{0.8cm}c}
\toprule
& & $x_0$ & $x_1$ & $x_2$ & $x_3$ & $x_4$ & $x_5$ & $x_6$ & $x_7$ & $x_8$ \\
& & \texttt{SOL} & you're & gonna & to & go & to & indeed & indeed & Amsterdam \\
\midrule
\texttt{SOL} & $y_0$ & \cellcolor{gray!32} \bf 0 & -1 & -2 & -3 & -4 & -5 & -6 & -7 & -8 \\
you're       & $y_1$ & -1 & \cellcolor{gray!32} \bf 2 & 1 & 0 & -1 & -2 & -3 & -4 & -5 \\
going        & $y_2$ & -2 & 1 & \cellcolor{gray!32} \bf 1 & 0 & -1 & -2 & -3 & -4 & -5 \\
to           & $y_3$ & -3 & 0 & 0 & \cellcolor{gray!32} \bf 3 & 2 & 1 & 0 & -1 & -2 \\
go           & $y_4$ & -4 & -1 & -1 & 2 & \cellcolor{gray!32} \bf 5 & 4 & 3 & 2 & 1 \\
to           & $y_5$ & -5 & -2 & -2 & 1 & 4 & \cellcolor{gray!32} \bf 7 & 6 & 5 & 4 \\
uh           & $y_6$ & -6 & -3 & -3 & 0 & 3 & 6 & 6 & 5 & 4 \\
Amsterdam    & $y_7$ & -7 & -4 & -4 & -1 & 2 & 5 & 5 & 5 & 7 \\
\bottomrule
\end{tabular}}
\caption{For $z_0 = \texttt{SOL}$.}
\label{tab:name-1}
\end{subtable}
\vspace{0.5em}

\begin{subtable}{\textwidth}
\centering\small{ 
\begin{tabular}{cc|>{\centering\arraybackslash}p{0.8cm}>{\centering\arraybackslash}p{0.8cm}>{\centering\arraybackslash}p{0.8cm}>{\centering\arraybackslash}p{0.8cm}>{\centering\arraybackslash}p{0.8cm}>{\centering\arraybackslash}p{0.8cm}>{\centering\arraybackslash}p{0.8cm}>{\centering\arraybackslash}p{0.8cm}c} 
\toprule
& & \texttt{SOL} & you're & gonna & to & go & to & indeed & indeed & Amsterdam \\
& & $x_0$ & $x_1$ & $x_2$ & $x_3$ & $x_4$ & $x_5$ & $x_6$ & $x_7$ & $x_8$ \\
\midrule
\texttt{SOL} & $y_0$ & -1 & -1 & -2 & -3 & -4 & -5 & -3 & -4 & -5 \\
you're       & $y_1$ & -2 & 1 & 1 & 0 & -1 & -2 & 0 & -1 & -2 \\
going        & $y_2$ & -3 & 0 & 0 & 0 & -1 & -2 & 0 & -1 & -2 \\
to           & $y_3$ & -4 & -1 & -1 & 2 & 2 & 1 & 3 & 2 & 1 \\
go           & $y_4$ & -5 & -2 & -2 & 1 & 4 & 4 & 6 & 5 & 4 \\
to           & $y_5$ & -6 & -3 & -3 & 0 & 3 & 6 & \cellcolor{gray!32} \bf 9 & 8 & 7 \\
uh           & $y_6$ & -7 & -4 & -4 & -1 & 2 & 5 & 8 & 8 & 7 \\
Amsterdam    & $y_7$ & -8 & -5 & -5 & -2 & 1 & 4 & 7 & 7 & 10 \\
\bottomrule
\end{tabular}}
\caption{For $z_1 = \textit{indeed}$.}
\label{tab:name-2}
\end{subtable}
\vspace{0.5em}

\begin{subtable}{\textwidth}
\centering\small{ 
\begin{tabular}{cc|>{\centering\arraybackslash}p{0.8cm}>{\centering\arraybackslash}p{0.8cm}>{\centering\arraybackslash}p{0.8cm}>{\centering\arraybackslash}p{0.8cm}>{\centering\arraybackslash}p{0.8cm}>{\centering\arraybackslash}p{0.8cm}>{\centering\arraybackslash}p{0.8cm}>{\centering\arraybackslash}p{0.8cm}c} 
\toprule
& & \texttt{SOL} & you're & gonna & to & go & to & indeed & indeed & Amsterdam \\
& & $x_0$ & $x_1$ & $x_2$ & $x_3$ & $x_4$ & $x_5$ & $x_6$ & $x_7$ & $x_8$ \\
\midrule
\texttt{SOL} &$y_0$ & -2 & -2 & -2 & -3 & -4 & -5 & -3 & -1 & -2 \\
you're       &$y_1$ & -3 & 0 & 0 & 0 & -1 & -2 & 0 & 2 & 1 \\
going        &$y_2$ & -4 & -1 & -1 & -1 & -1 & -2 & 0 & 2 & 1 \\
to           &$y_3$ & -5 & -2 & -2 & 1 & 1 & 1 & 3 & 5 & 4 \\
go           &$y_4$ & -6 & -3 & -3 & 0 & 3 & 3 & 6 & 8 & 7 \\
to           &$y_5$ & -7 & -4 & -4 & -1 & 2 & 5 & 8 & \cellcolor{gray!32} \bf 11 & 10 \\
uh           &$y_6$ & -8 & -5 & -5 & -2 & 1 & 4 & 7 & \cellcolor{gray!32} \bf 10 & 10 \\
Amsterdam    &$y_7$ & -9 & -6 & -6 & -3 & 0 & 3 & 6 & 9 & \cellcolor{gray!32} \bf 12 \\
\bottomrule
\end{tabular}}
\caption{For $z_2 = \textit{indeed}$.}
\label{tab:name-3}
\end{subtable}

\caption{The scoring matrix $F \in \mathbb{R}^{9 \times 8 \times 3}$ generated by Algorithm~\ref{alg:MSAwithPermutations} for the working example in Section~\ref{sec:align4d}. The third dimension of $F$ is represented by the three Sub-tables \ref{tab:name-1}, \ref{tab:name-2}, and \ref{tab:name-3}.}
\label{tab:scoring-matrix-3d}
\end{table*}

\begin{table*}[htbp!]
\centering\small{ 
\begin{tabular}{l|ccccccccc} 
\toprule
\bf $\bm{E}$   & $\bm{(x_1, y_1)}$ & $\bm{(x_2, y_2)}$ & $\bm{(x_3, y_3)}$ & $\bm{(x_4, y_4)}$ & $\bm{(x_5, y_5)}$ & $\bm{(x_6, z_1)}$ & $\bm{(x_7, z_2)}$ & $\bm{(y_6)}$ & $\bm{(x_8, y_7)}$ \\
\midrule
$X$ & You're & gonna & to & go & to & indeed & indeed & - & Amsterdam \\
$Y$ & You're & going & to & go & to & - & - & uh & Amsterdam \\
$Z$ & - & - & - & - & - & indeed & indeed & - & - \\
\bottomrule
\end{tabular}}
\caption{The alignment matrix produced by Algorithm~\ref{alg:backtracking} using the scoring matrix in Table~\ref{tab:scoring-matrix-3d}.}
\label{tab:alignment-matrix}
\end{table*}

\end{document}